\documentclass[sigconf]{acmart}
\graphicspath{ {images/} }
\usepackage{booktabs}

\begin{document}
\title{COEGAN: Evaluating the Coevolution Effect in Generative Adversarial Networks}

\author{Victor Costa}
\affiliation{%
	\institution{CISUC, Department of Informatics Engineering \\ University of Coimbra}
	\city{Coimbra}
	\country{Portugal}
}
\email{vfc@dei.uc.pt}

\author{Nuno Louren\c{c}o}
\affiliation{%
	\institution{CISUC, Department of Informatics Engineering \\ University of Coimbra}
	\city{Coimbra}
	\country{Portugal}
}
\email{naml@dei.uc.pt}

\author{Jo\~{a}o Correia}
\affiliation{%
	\institution{CISUC, Department of Informatics Engineering \\ University of Coimbra}
	\city{Coimbra}
	\country{Portugal}
}
\email{jncor@dei.uc.pt}

\author{Penousal Machado}
\affiliation{%
	\institution{CISUC, Department of Informatics Engineering \\ University of Coimbra}
	\city{Coimbra}
	\country{Portugal}
}
\email{machado@dei.uc.pt}

\begin{abstract}
Generative adversarial networks (GAN) present state-of-the-art results in the generation of samples following the distribution of the input dataset. However, GANs are difficult to train, and several aspects of the model should be previously designed by hand. Neuroevolution is a well-known technique used to provide the automatic design of network architectures which was recently expanded to deep neural networks.

COEGAN is a model that uses neuroevolution and coevolution in the GAN training algorithm to provide a more stable training method and the automatic design of neural network architectures. COEGAN makes use of the adversarial aspect of the GAN components to implement coevolutionary strategies in the training algorithm. Our proposal was evaluated in the Fashion-MNIST and MNIST dataset. We compare our results with a baseline based on DCGAN and also with results from a random search algorithm. We show that our method is able to discover efficient architectures in the Fashion-MNIST and MNIST datasets. The results also suggest that COEGAN can be used as a training algorithm for GANs to avoid common issues, such as the mode collapse problem.
\end{abstract}

\copyrightyear{2019}
\acmYear{2019}
\acmConference[GECCO '19]{Genetic and Evolutionary Computation Conference}{July 13--17, 2019}{Prague, Czech Republic}
\acmBooktitle{Genetic and Evolutionary Computation Conference (GECCO '19), July 13--17, 2019, Prague, Czech Republic}
\acmDOI{10.1145/3321707.3321746}

\settopmatter{printacmref=false}
\setcopyright{none}
\renewcommand\footnotetextcopyrightpermission[1]{}
\pagestyle{plain}



\maketitle

\section{Introduction}
Generative adversarial networks (GAN) \cite{NIPS2014_5423} gained relevance for presenting impressive results, mainly for image synthesis in the field of computer vision.
A GAN combines two components, a discriminator and a generator, trained as adversaries in an algorithm designed to minimize a previously defined cost function.
The generative component is trained without the direct awareness from the data distribution they are trying to capture.
Hence, the discriminator learns to distinguish between fake samples and the real input data and the generator learns to synthesize samples based on the input dataset.

Recently, GANs were improved to generate high-resolution images in large-scale datasets \cite{karras2018progressive,brock2018large}.
However, there are still open problems regarding the training of GANs.
Vanishing gradient and the mode collapse are the most common issues, making the training of GANs hard.
There are strategies to minimize these problems, but they remain fundamentally unsolved \cite{gulrajani2017improved,salimans2016improved}.

In a GAN, the discriminator and the generator are deep neural networks that should have architectures defined previously.
In this case, the topology and hyperparameters are usually empirically chosen, thus spending human time in repetitive tasks such as fine-tuning.
However, there are approaches that can automatize the design of neural network architectures.

Neuroevolution is a technique that applies evolutionary algorithms to provide the automatic design of neural networks.
In neuroevolution, both the network architecture (e.g., topology, hyperparameters and the optimization method) and the parameters (e.g., weights) can be evolved.
NeuroEvolution of Augmenting Topologies (NEAT) \cite{neat} is a well-known algorithm that evolves the weights and topologies of neural networks.
NEAT was also successfully applied in a coevolution context \cite{stanley2004competitive}.
The NEAT model was also expanded to work on larger search spaces, such as deep neural networks, in the DeepNEAT~\cite{miikkulainen2017evolving} method.

This paper presents a model called coevolutionary generative adversarial networks (COEGAN), first proposed in \cite{costa2019coegan}, that combines neuroevolution and coevolution in the coordination of the GAN training algorithm.
Our evolutionary algorithm is based on the approach used on DeepNEAT.
We extended and adapted DeepNEAT to work on the context of GANs, making use of the competitive characteristic between the generator and discriminator to apply a coevolution model.
Hence, each subpopulation of generators and discriminators evolve following its own evolutionary path.
To validate our model, experiments were conducted using MNIST \cite{lecun1998mnist} and Fashion-MNIST \cite{xiao2017fashion} as input datasets for the discriminator component.
We show that our model is better than a random search to discover architectures.
A comparison was made of our model with a reference architecture based on DCGAN \cite{radford2015unsupervised}.
We also show that the training stability is improved and our results are better when compared with manually designed networks with similar power \footnote{Code available at \href{https://github.com/vfcosta/coegan}{https://github.com/vfcosta/coegan}.}.

The remainder of this paper is organized as follows:
Section \ref{sec:background} introduces the concepts of GANs and evolutionary algorithms;
Section \ref{sec:coegan} presents our approach used to evolve GANs;
Section \ref{sec:experiments} displays the experimental results using this approach;
finally, Section \ref{sec:conclusions} presents our conclusions.

\section{Background} \label{sec:background}

This section introduces the concepts of evolutionary algorithms and generative adversarial networks employed in this paper and presents works related to the proposed model.

\subsection{Evolutionary Algorithms} \label{sec:evolutionary_algorithms}

Evolutionary algorithms (EAs) are a family of algorithms inspired by biological evolution, simulating the evolutionary mechanism found in nature~\cite{sims1994evolving}.
There are several variations and applications related to evolutionary algorithms proposed to solve a diverse variation of problems.
In this context, neuroevolution was proposed to apply evolutionary algorithms to evolve neural networks~\cite{neat}.
Neuroevolution can be used to evolve weights, topology and hyperparameters of a neural network.
In this paper, we are particularly interested in the use of neuroevolution to automate the design of the network architecture and its parameters.
This automation is even more relevant for bigger models such as deep neural networks, which produces large search spaces \cite{assunccao2018evolving,miikkulainen2017evolving}.

NeuroEvolution of Augmenting Topologies (NEAT) is a well-known method used to evolve the topology and weights of neural networks.
NEAT encodes in the genotype the structure and weights of the neural network.
The genes represent neurons and connections between them, including the weights used in the transformation to the phenotype (i.e., the resulting neural network).
The evolution occurs through mutation and crossover.
The growth strategy follows a complexifying mechanism where the genome starts small and gradually grows with the generations.
In NEAT, not only the final architecture is important, but the intermediary solutions also contribute to the final solution, since the weights are transferred through generations \cite{neat}.
DeepNEAT \cite{miikkulainen2017evolving} was proposed to extend the NEAT model to deep neural networks.
In DeepNEAT, each gene in the genotype represents an entire layer of the neural network.
This approach makes it possible to discover deeper models.

Coevolution is the simultaneous evolution of at least two distinct species \cite{hillis1990co,rawal2010constructing}.
In \cite{stanley2004competitive}, NEAT was applied in a competitive coevolution environment.
In competitive coevolution, individuals of two or more species are competing between them.
Therefore, the fitness function represents the competition in order to represent a score that is inversely related between different species \cite{stanley2004competitive,sims1994evolving,rawal2010constructing}.

\subsection{Generative Adversarial Networks} \label{sec:gan}
Generative Adversarial Networks (GAN) \cite{NIPS2014_5423} became relevant for the performance achieved in generative tasks, mainly in the field of computer vision.
A GAN combines a discriminator $D$ and a generator $G$, trained simultaneously as adversaries, to create strong generators and discriminators.
The discriminator $D$ aims to distinguish between real data and fake samples, given an input distribution.
The generator $G$ has the objective to outputs fake samples to deceive the discriminator, capturing the data distribution used as input for $D$.

The loss function of the discriminator is defined as following:

\begin{equation}
J^{(D)}(D,G) = -\mathbb{E}_{x \sim p_{data}}[\log D(x)] - \mathbb{E}_{z \sim p_z}[\log(1 - D(G(z)))].
\label{eq:discriminator}
\end{equation}

For the generator, the non-saturating version of the loss function is defined by:

\begin{equation}
J^{(G)}(G) = - \mathbb{E}_{z \sim p_z}[\log(D(G(z)))].
\label{eq:generator}
\end{equation}

$p_{data}$ (Eq. \ref{eq:discriminator}) and $p_z$ (Eq. \ref{eq:generator}) represent the dataset used as input to the discriminator and the noise variable used as input to the generator, respectively.

Vanishing gradient and mode collapse are the most common problems regarding the training stability in GANs.
The vanishing gradient occurs when the discriminator $D$ became powerful enough to not be fooled by the generator anymore, avoiding the gradient to flow through the generator.
This causes the whole training progress to stagnate.
On the other hand, the mode collapse problem makes the generator to capture only a portion of the input distribution.
Several approaches tried to minimize those problems, but they remain unsolved \cite{salimans2016improved,gulrajani2017improved}.

Distinct variations of the original GAN were proposed in order to improve the stability and the performance of the model, such as WGAN \cite{arjovsky2017wasserstein} and LSGAN \cite{mao2017least}.
However, a study found no empirical evidence that those proposals are superior to the original GAN \cite{lucic2017gans}.

Other models propose an expansion to the original GAN proposal, modifying aspects of the training algorithm.
The method described by \cite{karras2018progressive} uses a progressive strategy to progressively grow a GAN during the training phase.
This mechanism increases the number of layers in the discriminator and generator as the training proceeds, augmenting the resolution of images at each phase.
The model proposed in \cite{karras2018progressive} evolves in a preconfigured way during the training, without the use of an evolutionary algorithm to guide this process.
Therefore, we can consider this predefined progression in the number of layers as a first step towards the evolution of generative adversarial models.

Recently, a model was proposed to use evolutionary algorithms in GANs \cite{wang2018evolutionary}.
Their approach used a simple model to evolve GANs, using a mutation operator that can switch only the loss function of the individuals.
They also use very small populations of individuals, only to capture the possibilities of losses predefined in the definition of the model.
Our proposal differs from them by modeling the GAN as a coevolution problem.
Besides, in our case, the evolution does not take into account the loss function and occurs only in the network architecture.
Coevolution was also used to train GANs in \cite{schmiedlechner2018towards}.
In this case, the training method focused on the evolution of weights and does not evolve the network architecture.

\section{COEGAN} \label{sec:coegan}
In \cite{costa2019coegan}, a model called Coevolutionary Generative Adversarial Networks (COEGAN) was proposed.
In this model, neuroevolution and coevolution are used in the coordination of the training algorithm.
So, COEGAN extended and adapted the approach of DeepNEAT \cite{miikkulainen2017evolving} to the context of GANs in a coevolution environment.

The genome of the COEGAN model is represented as an array of genes.
The phenotype transformation maps this array into a sequence of layers in a deep neural network.
Three types of genes are used in our method: linear, convolution and transpose convolution.
Each gene has an activation function, randomly chosen from the following set: ReLU, LeakyReLU, ELU, Sigmoid and Tanh.

The convolution and transpose convolution layers only have the number of output channels as a random parameter.
The stride and kernel size are dynamically calculated based on the requirements of the output size of each layer.
In addition, the number of input channels is also dynamically calculated, based on the previous layer.
The linear layer only has the number of output features as the random parameter.
The number of input features is defined based on the setup of the previous layer.
Therefore, genes belonging to a genome have only the activation function, output features and output channels subject to the mutation operation.

\begin{figure}
	\includegraphics[width=0.4\textwidth]{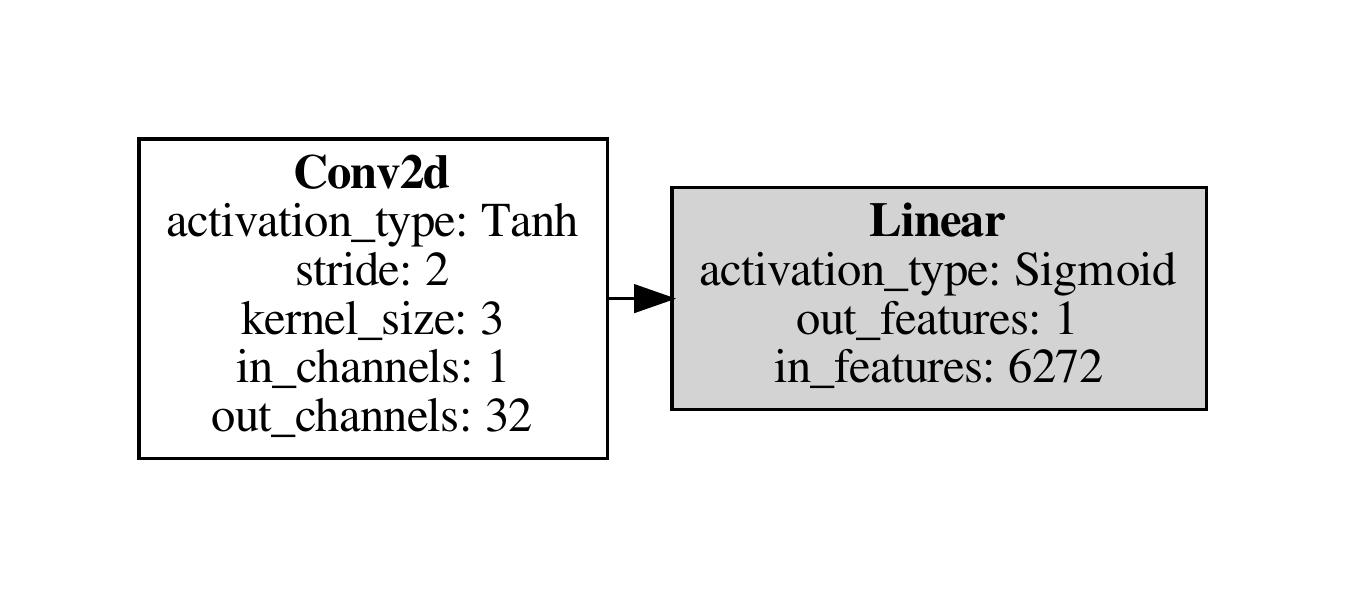}
	\caption{Example of a discriminator genotype}
	\label{fig:discriminator_genotype}
\end{figure}

\begin{figure}
	\includegraphics[width=0.4\textwidth]{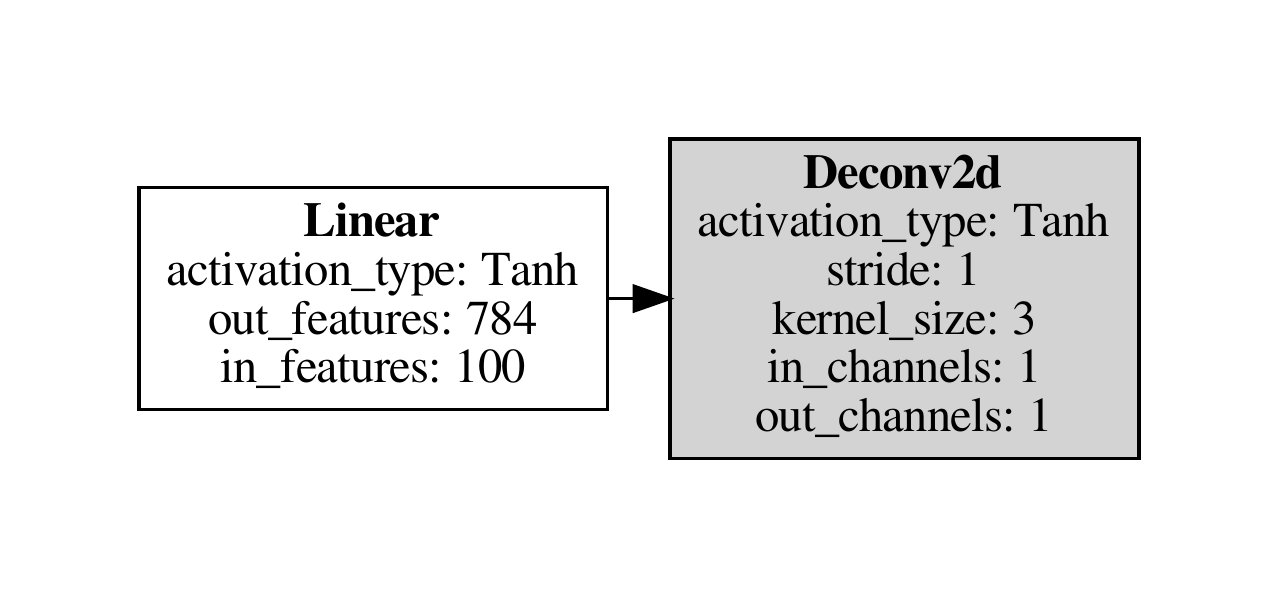}
	\caption{Example of a generator genotype}
	\label{fig:generator_genotype}
\end{figure}

Figure \ref{fig:discriminator_genotype} and Figure \ref{fig:generator_genotype} represent examples of the genotype of a discriminator and a generator, respectively.
In Figure \ref{fig:discriminator_genotype}, the genotype contains a convolutional section (defined by the Conv2d layer) and is followed by a linear section.
In Figure \ref{fig:generator_genotype}, the genotype contains a linear section followed by a transpose convolutional section (expressed by the Deconv2d layer).
Following the classical GAN approach, the output of discriminators is the probability of the input sample be a real sample, i.e., a sample drawn from the input dataset.
On the other hand, the generator output is a fake sample, with the same characteristics (i.e., dimension and channels) of a real sample.

In COEGAN, the population is composed of two subpopulations of generators $G_i$ and discriminators $D_i$.
Inside each subpopulation, a speciation mechanism inspired on the strategy used in NEAT is applied in order to promote innovation in each subpopulation.
This mechanism ensures that individuals with new layers will have the chance to survive long enough to be as powerful as individuals from previous generations.

For the current proposal of COEGAN, we are only interested in the evolution of the neural network architecture.
The number of parameters internal to the neural network, such as weights and bias, are too large and evolving them will increase the computational complexity.
Therefore, the parameters of the resulting neural networks will be trained by the gradient descent method and will not be part of the evolution.

\subsection{Fitness} \label{sec:fitness}

The fitness used in discriminators is based on the loss obtained from the regular GAN, i.e., the fitness is equivalent to Eq. \ref{eq:discriminator}.
For generators, preliminary experiments demonstrated that using the loss (Eq. \ref{eq:generator}) does not represent a good measure for quality in this case.
The loss for generators is not stable enough during the training, making it not suitable to be used as selection criteria in COEGAN.

Therefore, the Fr\'{e}chet Inception Distance (FID) \cite{heusel2017gans} was used in COEGAN as the fitness for generators.
FID is becoming the standard measurement to compare the performance of the generative component of GANs \cite{lucic2017gans}, having a better representation of diversity and quality than other metrics, such as the Inception Score \cite{salimans2016improved}.
Using the FID score we put selection pressure in generators and direct the evolution of the population towards strong generators with respect to this metric.
In FID, a hidden layer of Inception Net~\cite{szegedy2016rethinking} (trained on ImageNet~\cite{russakovsky2015imagenet}) is used in the transformation of images into the feature space, interpreted as a continuous multivariate Gaussian.
This transformation is applied to a subset of the real dataset and fake samples created by the generator.
The mean and covariance of the two resulting Gaussians are estimated and the Fr\'{e}chet distance between these Gaussians is given by:

\begin{equation}
FID(x,g) = ||\mu_x - \mu_g||_2^2 + Tr(\varSigma_x + \varSigma_g - 2(\varSigma_x\varSigma_g)^{1/2}).
\label{eq:fid}
\end{equation}

\subsection{Variation Operator} \label{sec:variation_operators}
The variation operator used in COEGAN to breed new individuals is the mutation operator.
We also experimented with a crossover operator, but the results indicated that this brings too much instability into the system.
Thus, we choose to keep only the mutation for the COEGAN proposal.

The mutation consists of the following kinds of operations: add a new layer, remove a layer, and change an existing layer.
The addition operator adds a new layer into the genotype.
This new layer is randomly drawn from a set of possible layers: linear and convolution for discriminators; linear and transpose convolution for generators.

The remove operation randomly chooses an existing layer and excludes it from the genotype.
On the other hand, the change operation modifies the attributes and the activation function of an existing layer.
In this case, the activation function is randomly chosen from the set of possibilities listed before.
Furthermore, specific attributes for layers can also be changed.
For the dense and convolution layers, the number of output features and the number of output channels can be mutated, respectively.
The mutation of these attributes follows a uniform distribution, delimited by a predefined range.

In the breeding process, the parameters (weights and bias) are copied when the genes involved in the mutation are compatible.
So, the new individual will keep the training information from the previous generation.
However, when the specific attributes of a linear or convolution layer change, the trained parameters are not copied and the layer will be trained from the beginning.
This is caused by the change in the shape of weights, making them incompatible with the new layer.

\subsection{Pairing Strategy} \label{sec:pairing}
In a competitive coevolution environment, discriminators and generators must be paired to calculate the fitness for individuals.
In this context, several approaches can be used to pair individuals, such as \textit{all vs. all}, \textit{random}, and \textit{all vs. best} \cite{sims1994evolving}.
The \textit{all vs. all} approach pairs each discriminator with each generator to calculate the fitness for each individual.
In this case, the fitness for discriminators will be the average of the losses obtained by each training pair.
The \textit{random} approach randomly pairs individuals from the discriminator and generator populations.
In the \textit{all vs. best} strategy, each individual in one population is paired with the best individual from the other population.

Preliminary experiments indicated that the \textit{all vs. all} strategy is the most stable for COEGAN.
This strategy improves the variability of the environment for both discriminators and generators during the training, helping to avoid common problems in the training of GANs.
The trade-off is the complexity of this strategy in respect to the execution time.

\subsection{Selection} \label{sec:selection}
The selection phase of COEGAN is based on the original proposal of NEAT~\cite{neat}.
The population of generators and discriminators are divided into species, which contains individuals with similar network structures.
The similarity criterion between individuals is based on the parameters of the genome that are related to the evolutionary algorithm.
Thus, we do not consider the weights and bias of each layer in this calculation. 

The criterion is represented by the distance $\delta$ between two genomes $i$ and $j$, defined as the number of genes that exist exclusively in $i$ or $j$.
The species are grouped based on the distance $\delta$ and a threshold $\delta_t$.
The $\delta_t$ parameter is adjusted automatically by the COEGAN algorithm to fit a predefined number of species.
Tournament was also applied to select the best individuals inside each species.

\section{Experiments} \label{sec:experiments}
To validate the performance of our method, we experiment COEGAN with the MNIST \cite{lecun1998mnist} and Fashion-MNIST \cite{xiao2017fashion} datasets.
We evaluate COEGAN against a random search method and a reference architecture based on DCGAN.
The random search method is similar to COEGAN, but instead of the fitness described in Section \ref{sec:fitness}, we use a random method to represent the fitness of individuals in the population.
All other characteristics of the random method, such as the pairing strategy, remain the same as used in COEGAN.
The DCGAN model is a well-defined set of architectural constraints, used as reference in several works related to evaluations of GANs \cite{lucic2017gans, karras2018progressive}.
We follow this approach to build a reference architecture (based on DCGAN) to compare our results with commonly used models in the context of GANs.

\subsection{Experimental Setup}
Table \ref{table:setup} presents the parameters used in our experiments.
The number of generations used in all experiments is $50$.
We used 10 individuals for each population of generators and discriminators.
These evolutionary parameters do not apply to the DCGAN experiments as DCGAN is not trained by an evolutionary algorithm.
In order to limit the computational resources used in our experiments, the size of the genome was limited to six layers.
To emulate a network with similar power, the DCGAN architecture used in the experiments also contains six layers.
We use three species for each population of generators and discriminators, which allow an average of $3$ individuals per species.
We empirically defined a probability of 20\%, 10\% and 10\% for the add, remove and change mutations, respectively.
A higher probability for these mutations causes the premature convergence of the system, leading to performance issues and instability on the GAN training process.
Hence, the probability rates were kept low but sufficient to create diversity in the population through generations.

For each pair of $(G_i$, $D_i)$, $20$ batches were executed per generation, with the batch size of $64$.
Therefore, in our scenario of a population composed of $10$ generators and $10$ discriminators with the \textit{all vs. all} pairing strategy, each individual will execute $200$ batches per generation.
The DCGAN experiment is not an evolutionary algorithm and contains only one discriminator and one generator.
In this case, we set the number of batches to $200$ to keep it comparable with COEGAN and the random search method.
The optimizer used in the training method was Adam~\cite{kingma2015adam} with a learning rate of $0.001$.

\begin{table}
	\caption{Experimental parameters.}
	\begin{center}\begin{tabular}{c|c}
			\textbf{Evolutionary Parameters} & \textbf{Value} \\
			\hline
			Number of generations & 50 \\
			Population size (generators) & 10 \\
			Population size (discriminators) & 10 \\
			Add Layer rate & 20\% \\
			Remove Layer rate & 10\% \\
			Change Layer rate & 10\% \\
			Output features range & [32, 1024] \\
			Output channels range & [16, 128] \\
			Tournament $k_t$ & 2 \\
			FID samples & 1000 \\
			Root mean squared error samples & 1000 \\
			Genome Limit & 6 \\
			Species & 3 \\
			\textbf{GAN Parameters} & \textbf{Value} \\
			\hline
			Batch size & 64 \\
			Batches per generation & 20 \\
			Optimizer & Adam \\
			Learning rate & 0.001
	\end{tabular}\end{center}
	\label{table:setup}
\end{table}

\subsection{Results}

To compare the results of COEGAN, the random search method and the DCGAN based network, we use the FID score \cite{heusel2017gans}, Inception score \cite{salimans2016improved} and the root mean squared error.
The root mean squared error is calculated between samples created by the generator and real samples randomly drawn from the input dataset.

All figures in this section contain plots with curves representing the average of the results from $10$ repeated executions, with a confidence interval of $95\%$.

\subsubsection{MNIST} \label{sec:results_mnist}

\begin{figure}[h]
	\includegraphics[width=0.4\textwidth]{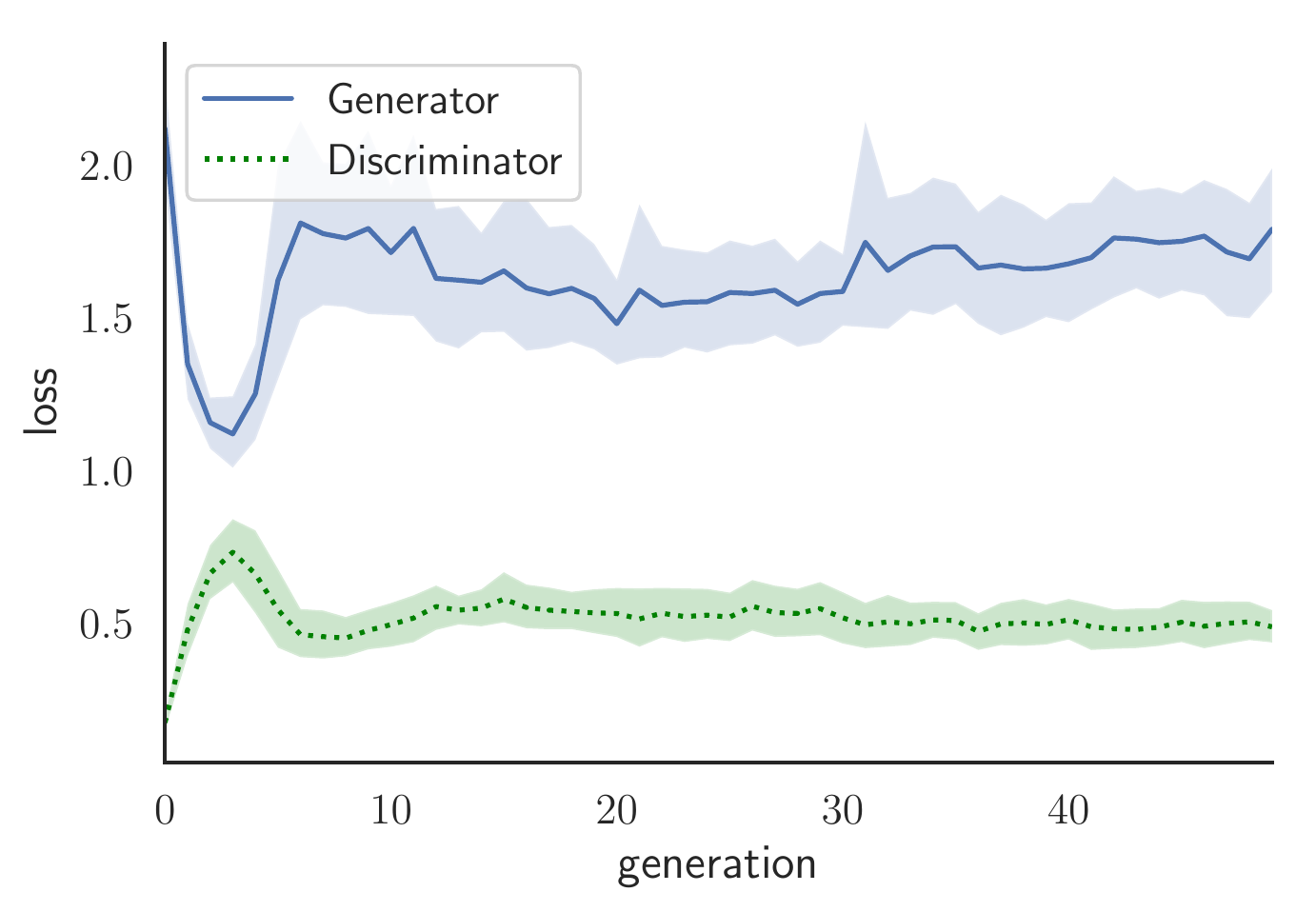}
	\caption{Losses of the discriminator and generator on the MNIST dataset.}
	\label{fig:mnist_loss_d}
\end{figure}

Figure \ref{fig:mnist_loss_d} displays the average of losses for the best generator and discriminator found for COEGAN in each generation.
As stated in Section \ref{sec:fitness}, this figure indicates that the use of the loss function as the fitness for generators is not a good metric to assess the performance of an individual.
We can see the value of the loss increases with generations as well as some instability in the values.

\begin{figure}[h]
	\includegraphics[width=0.4\textwidth]{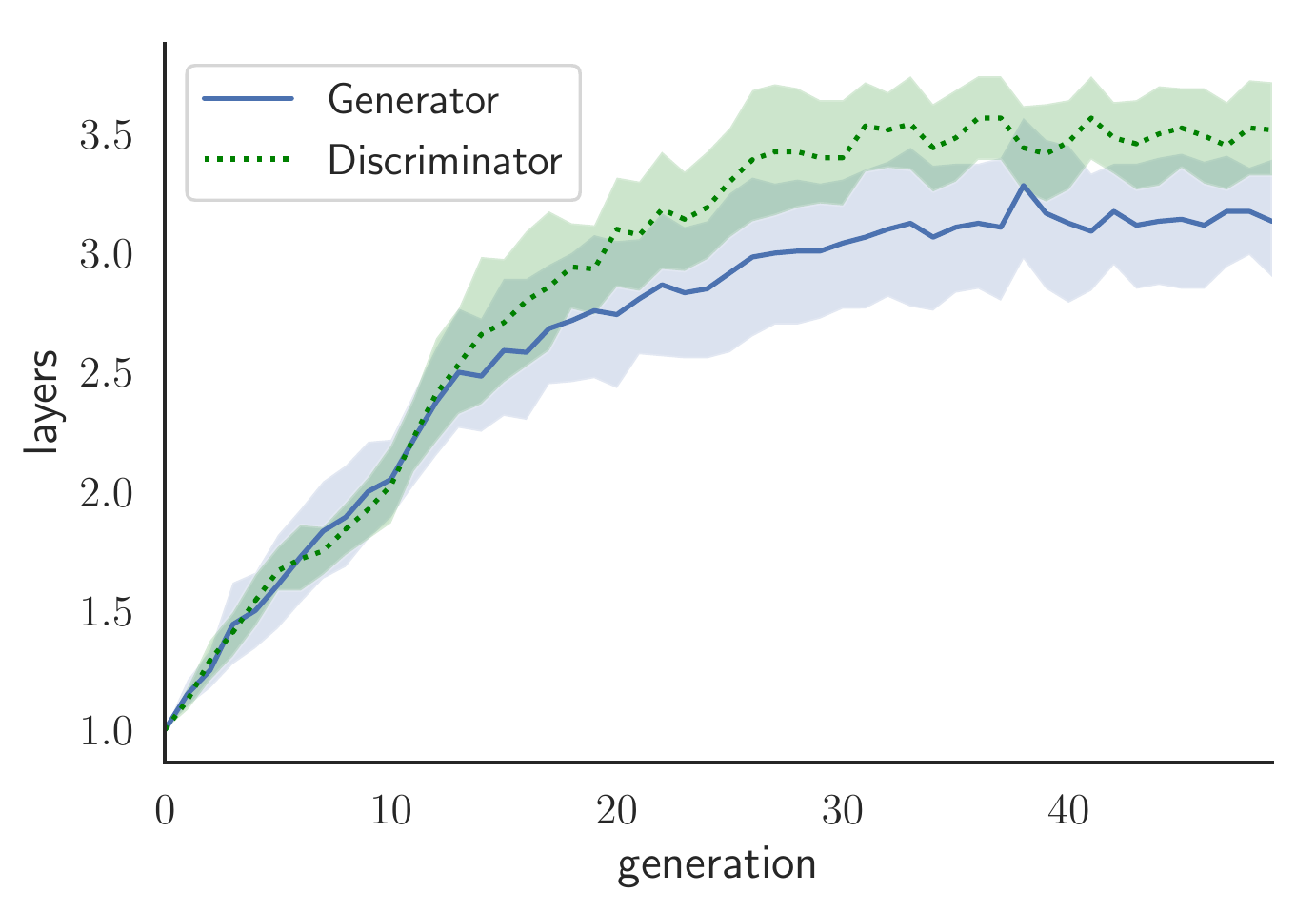}
	\caption{The average number of layers on the MNIST dataset.}
	\label{fig:mnist_layers_d}
\end{figure}

Figure \ref{fig:mnist_layers_d} presents the average progression of layers in the genome of individuals belonging to the population of generators and discriminators.
The number of layers gradually increases with generations, demonstrating that the speciation mechanism used in COEGAN protects the innovation and creates a propitious environment for individuals with more layers.

\begin{figure}[h]
	\includegraphics[width=0.4\textwidth]{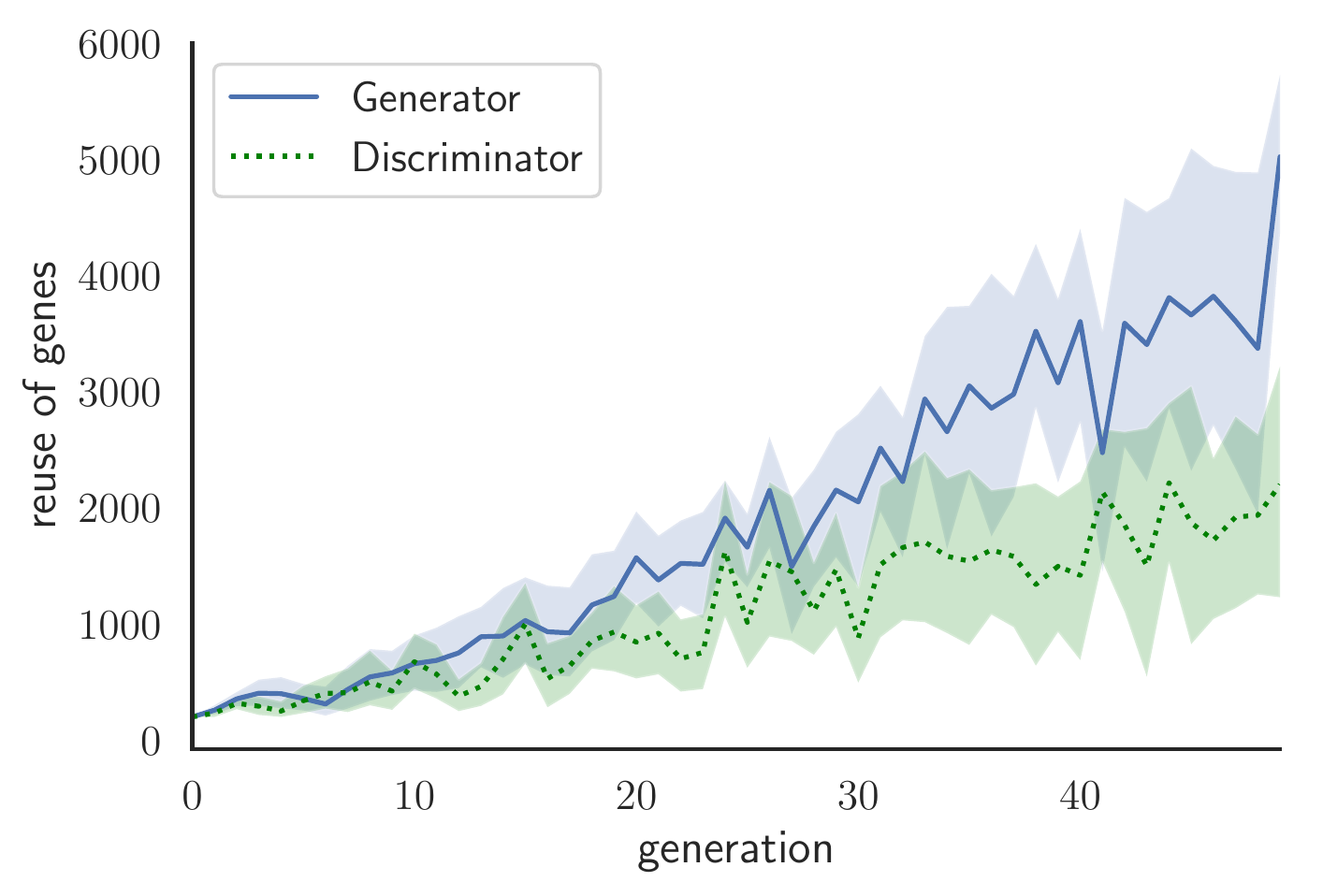}
	\caption{The average number of times a gene was reused when trained on the MNIST dataset.}
	\label{fig:mnist_genes_used_d}
\end{figure}

Figure \ref{fig:mnist_genes_used_d} displays the average number of times a gene was reused during the training process on the MNIST dataset.
The results for this metric proves the information is kept through generations described in Section \ref{sec:variation_operators}.

\begin{figure}[h]
	\includegraphics[width=0.4\textwidth]{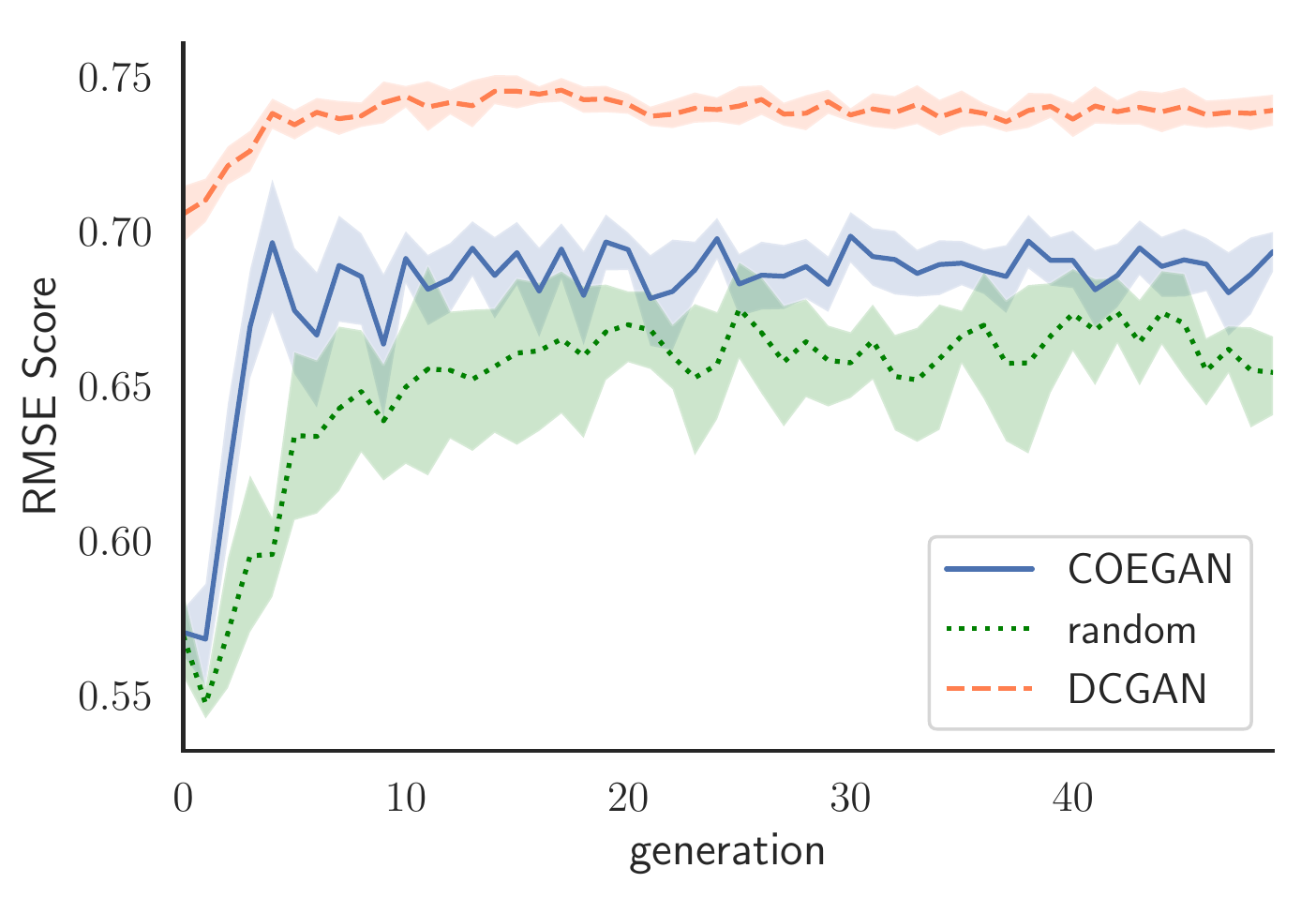}
	\caption{Root mean squared error on the MNIST dataset.}
	\label{fig:mnist_rmse_score_g}
\end{figure}

The root mean squared error is displayed in Figure \ref{fig:mnist_rmse_score_g}, comparing the results for COEGAN, the random method and DCGAN.
We introduced this metric to ensure that the samples created by generators are different from the data contained in the input dataset.
Thus, Figure \ref{fig:mnist_rmse_score_g} indicates that all methods create some innovation in the new samples, with the DCGAN method being better for this metric.

\begin{figure}[h]
	\includegraphics[width=0.4\textwidth]{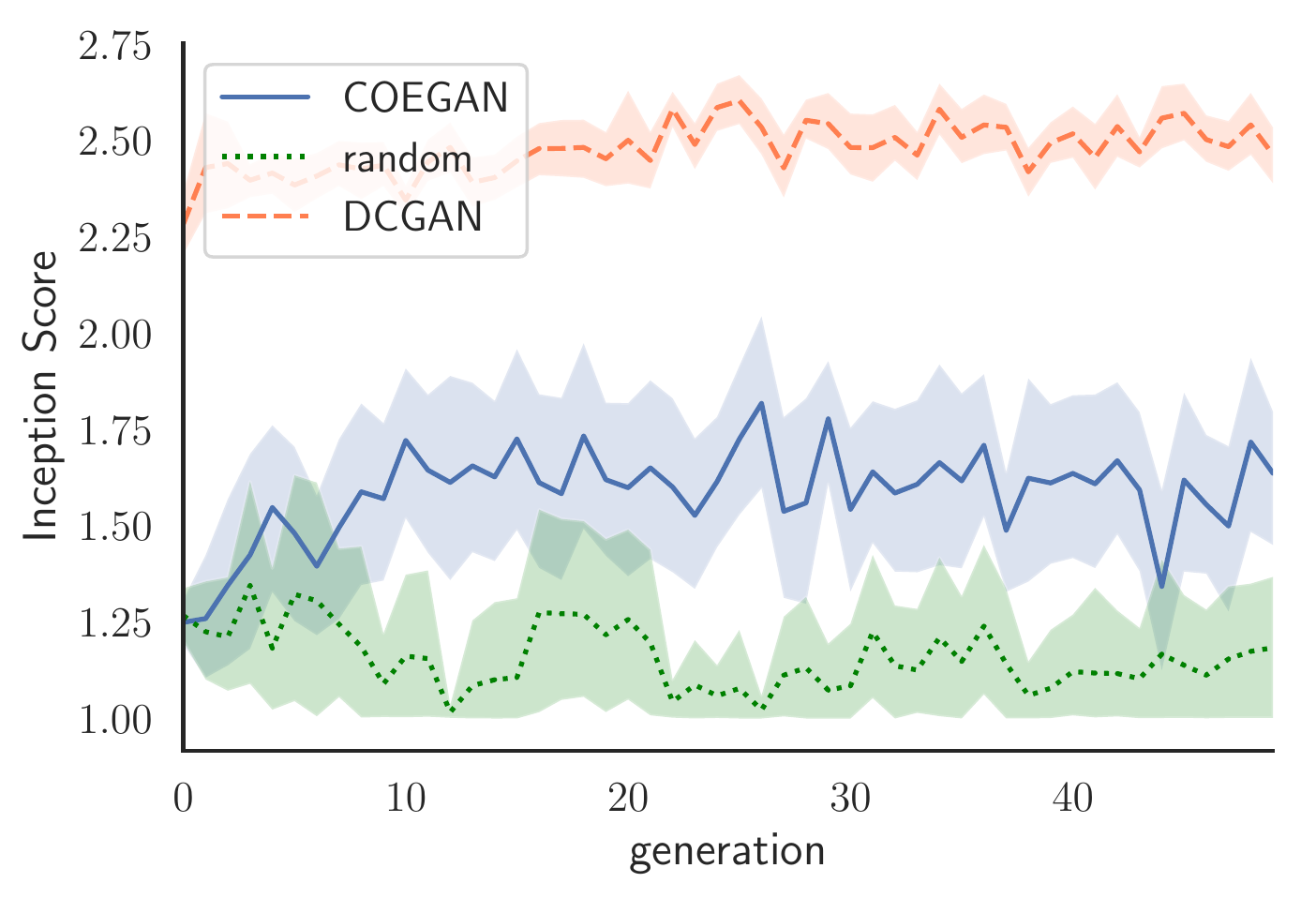}
	\caption{Inception Score on the MNIST dataset.}
	\label{fig:mnist_inception_score_g}
\end{figure}

Figure \ref{fig:mnist_inception_score_g} shows the average of the Inception Score (higher is better) for generators in COEGAN, the random method and DCGAN.
For this metric, the DCGAN provides the best results.
However, COEGAN is better than the random approach, demonstrating that our choice for fitness is relevant for the evolutionary algorithm proposed in this paper.

\begin{figure}[h]
	\includegraphics[width=0.4\textwidth]{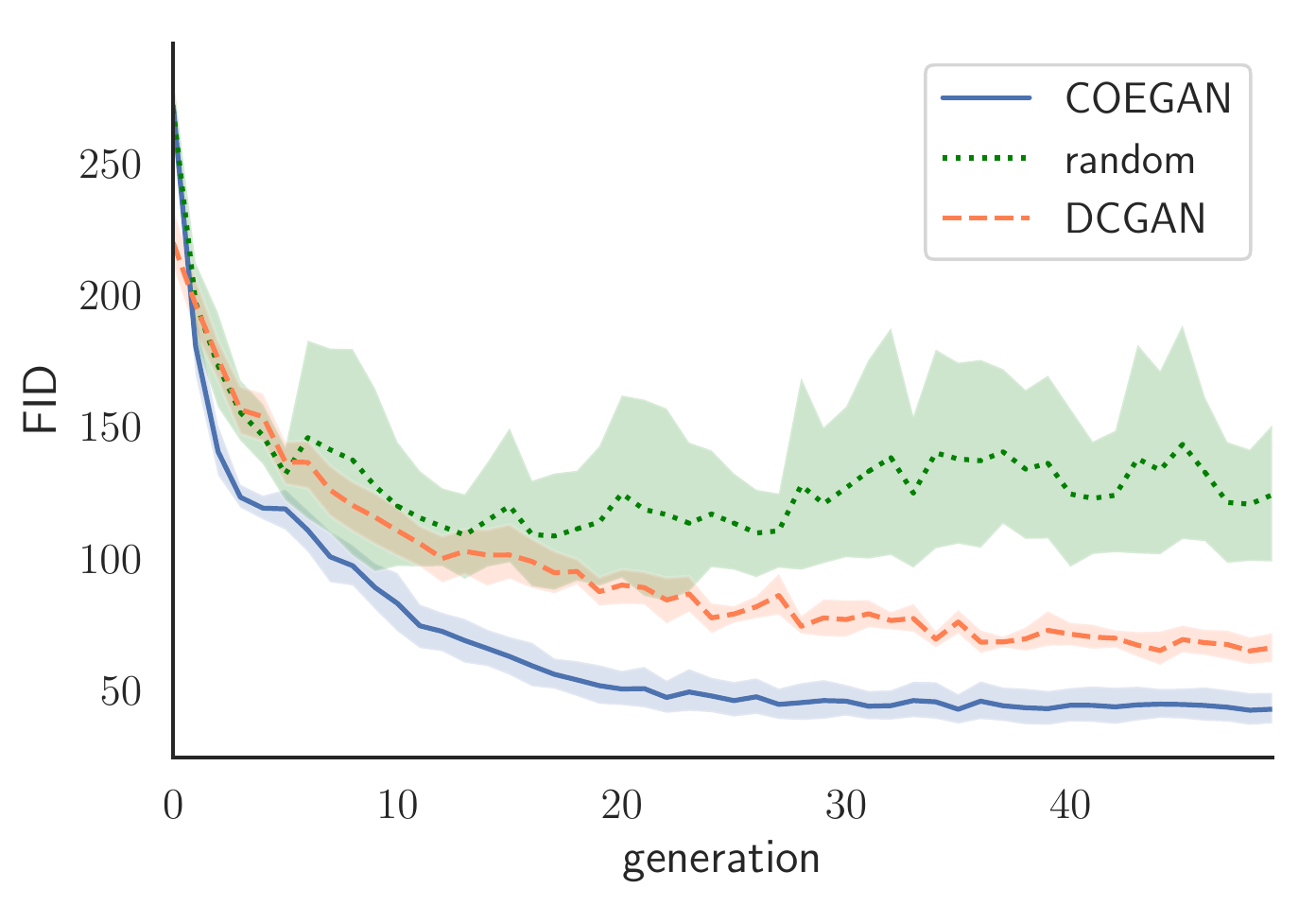}
	\caption{FID Score on the MNIST dataset.}
	\label{fig:mnist_fid_score_g}
\end{figure}

The Fr\'{e}chet Inception Distance (FID) \cite{heusel2017gans} of the generators in COEGAN, the random method and DCGAN are displayed in Figure \ref{fig:mnist_fid_score_g} (lower is better).
We can see that the FID for COEGAN is better than the results of the random method and DCGAN.
Moreover, the random method displayed a lot of variability in the FID results, mainly caused by the stochastic process introduced by this approach.
The study made in \cite{lucic2017gans} found that the FID score is a better representation of diversity and quality of generated samples when compared to real samples.
Thus, based on this study and the results displayed in Figure  \ref{fig:mnist_fid_score_g}, the best generator found in COEGAN outperforms the generator in the reference architecture based on DCGAN.

\begin{figure}[h]
	\includegraphics[width=0.4\textwidth]{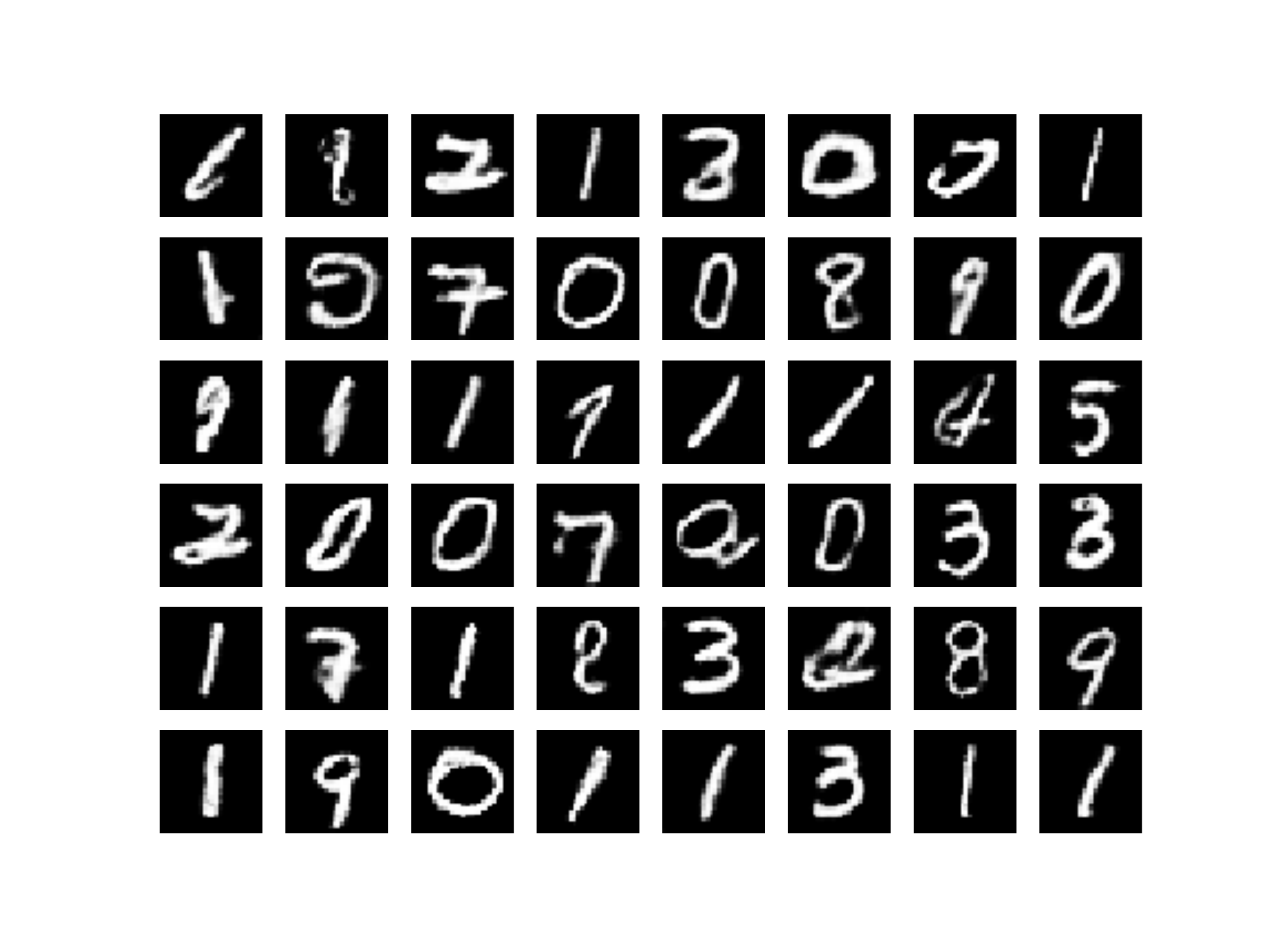}
	\caption{Samples generated by COEGAN when training on the MNIST dataset.}
	\label{fig:mnist_samples}
\end{figure}

Figure \ref{fig:mnist_samples} contains samples generated by the best generator found in COEGAN trained with the MNIST dataset after $50$ generations.
We can observe a good representation of the MNIST dataset in the generated samples.
We found no evidence of the vanishing gradient and the mode collapse problem in all executions of COEGAN.
As individuals with these issues perform worse than others, they will eventually not be selected by the evolutionary algorithm, preventing these problems to persist through generations.
Furthermore, a diverse population of generators and discriminators can increase the variability provided in the training process when compared to a regular GAN.
This variation contributes to a stronger training algorithm, preventing the mode collapse and the vanishing gradient problems.

\begin{figure}[h]
	\includegraphics[width=0.3\textwidth]{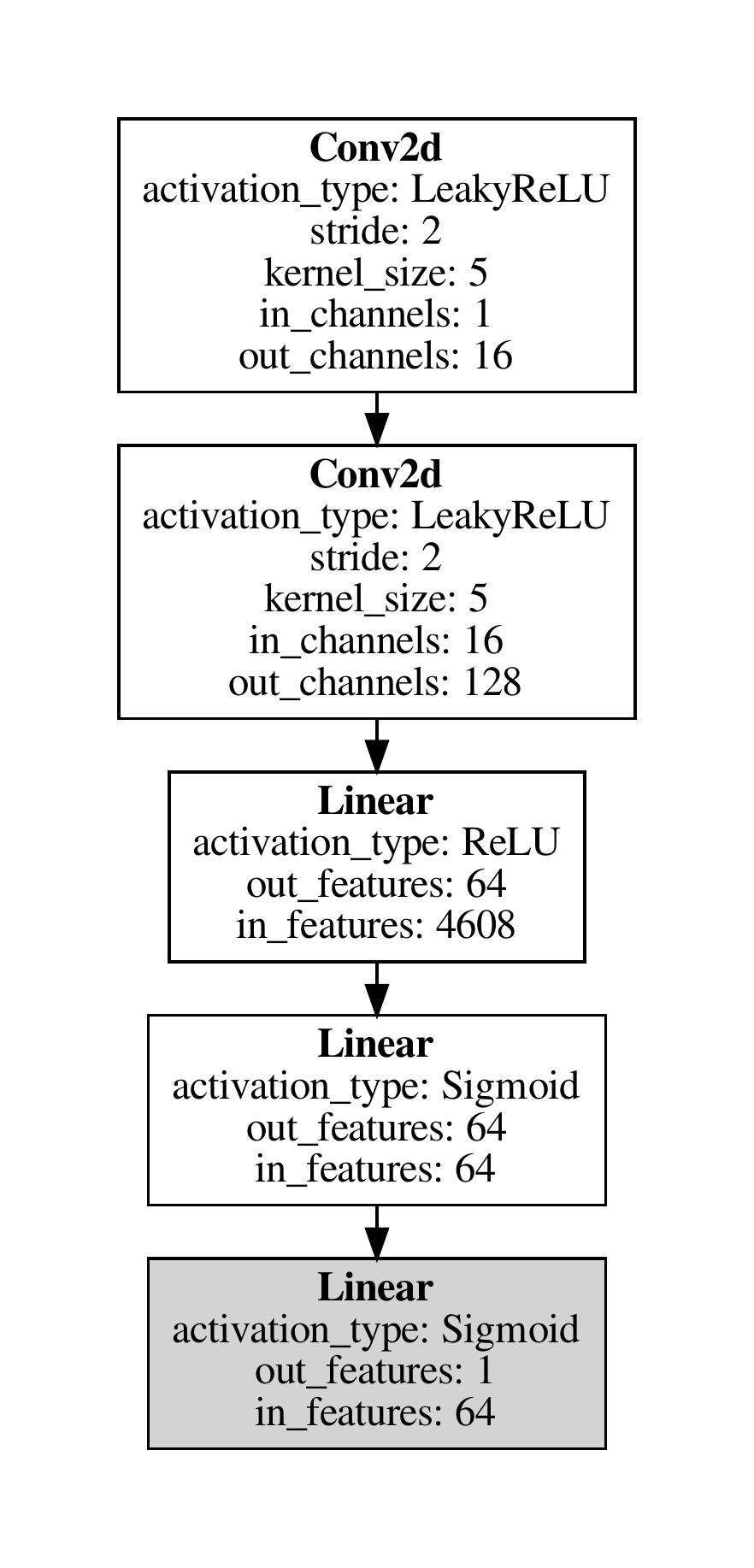}
	\caption{Best discriminator found by COEGAN when training on the MNIST dataset.}
	\label{fig:best_discriminator}
\end{figure}

\begin{figure}[h]
	\includegraphics[width=0.3\textwidth]{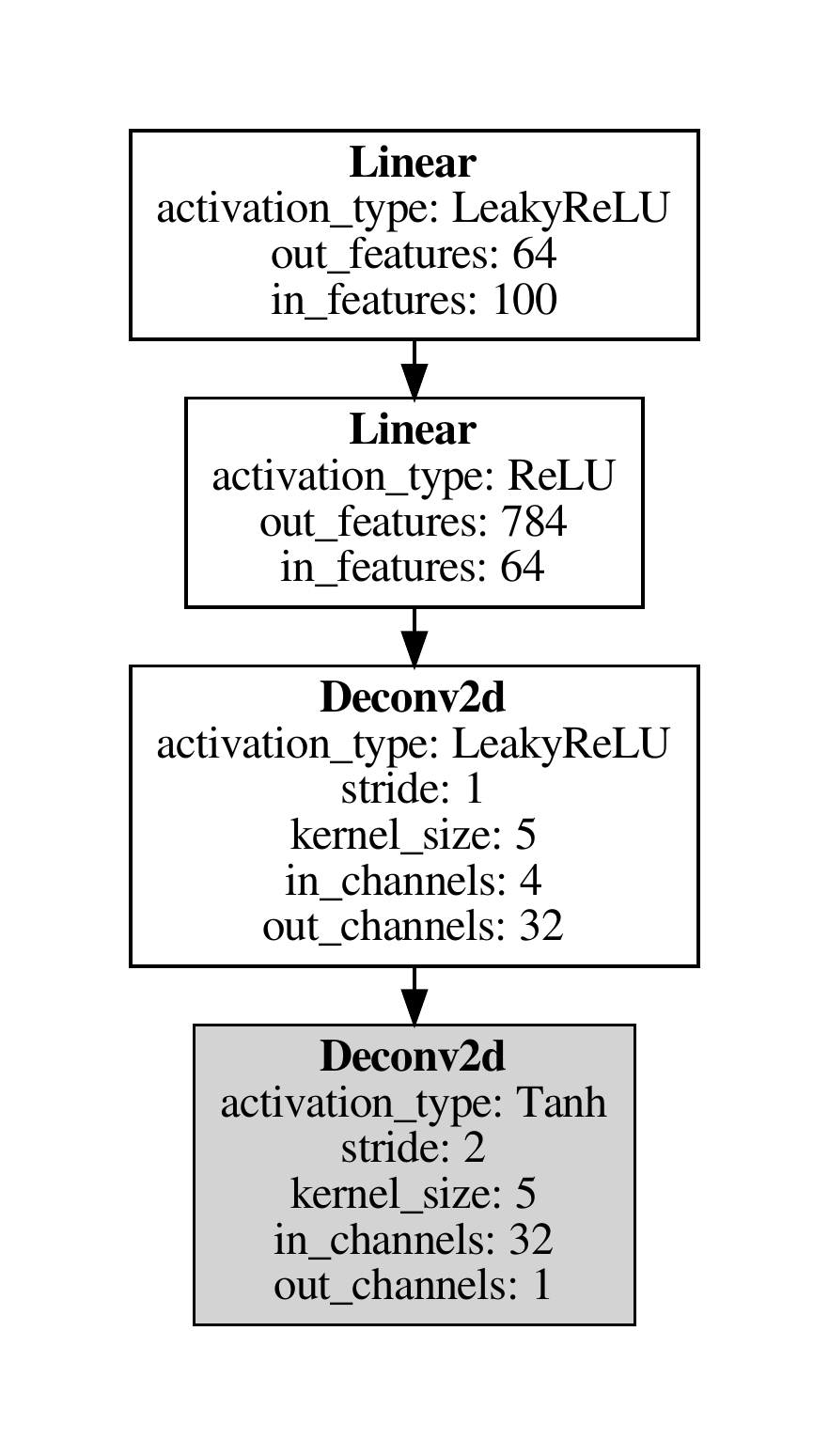}
	\caption{Best generator found by COEGAN when training on the MNIST dataset.}
	\label{fig:best_generator}
\end{figure}

Figures \ref{fig:best_discriminator} and \ref{fig:best_generator} represent the best architecture found by COEGAN after $50$ generations.
Both architectures are composed by a combination of linear and convolutional layers (represented in the images by Conv2d and Deconv2d).
It is relevant to note that not only the final architecture is important but also the process to construct the final models because of the mechanism of transference of the learned weights through generations.
Therefore, COEGAN found models for the generator and the discriminator with less layers than the reference architecture based on DCGAN, but with better performance with respect to the FID metric.

\subsubsection{Fashion-MNIST}
\begin{figure}[h]
	\includegraphics[width=0.4\textwidth]{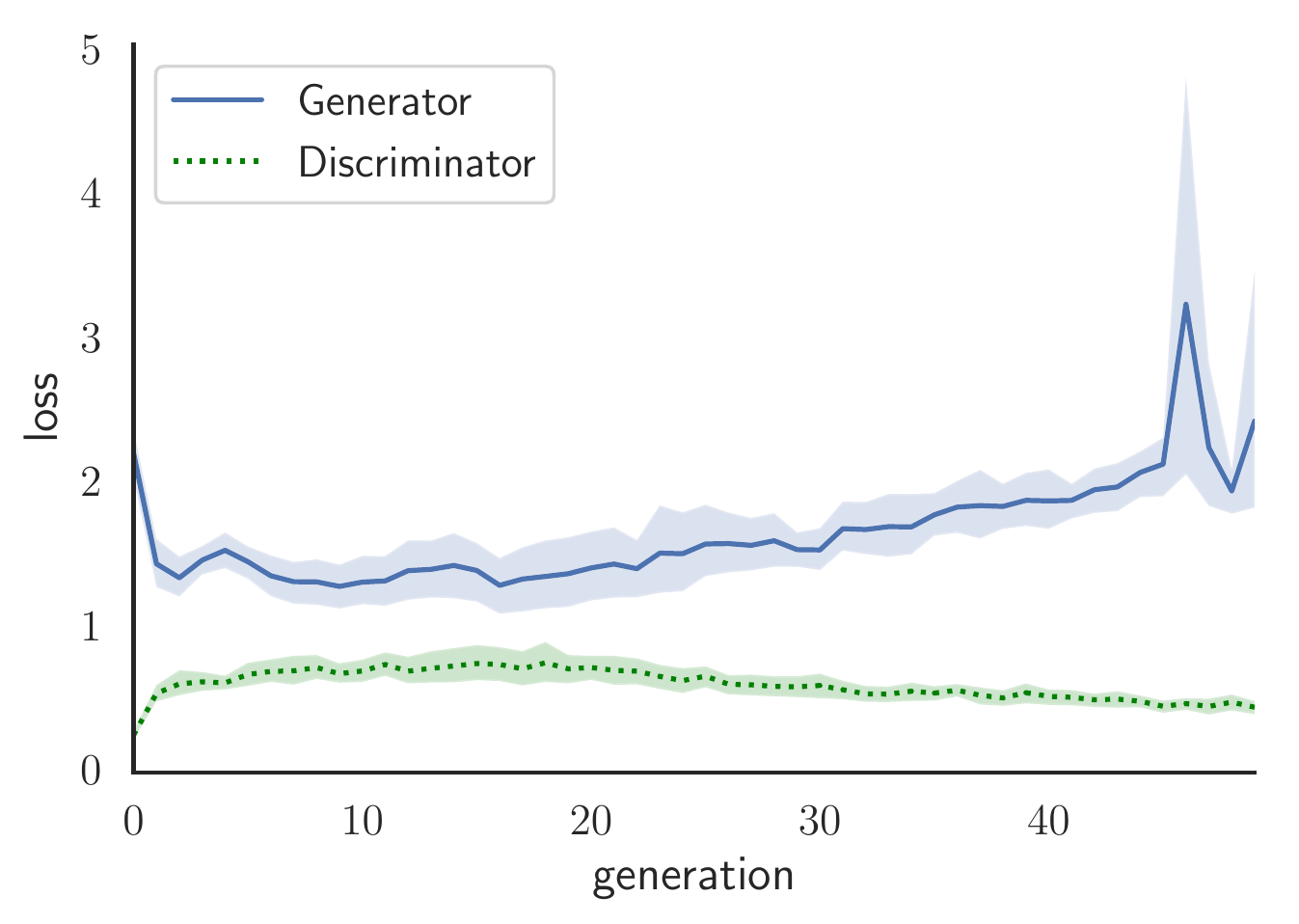}
	\caption{Losses of the discriminator and generator on the Fashion-MNIST dataset.}
	\label{fig:fashionmnist_loss_d}
\end{figure}

\begin{figure}[h]
	\includegraphics[width=0.4\textwidth]{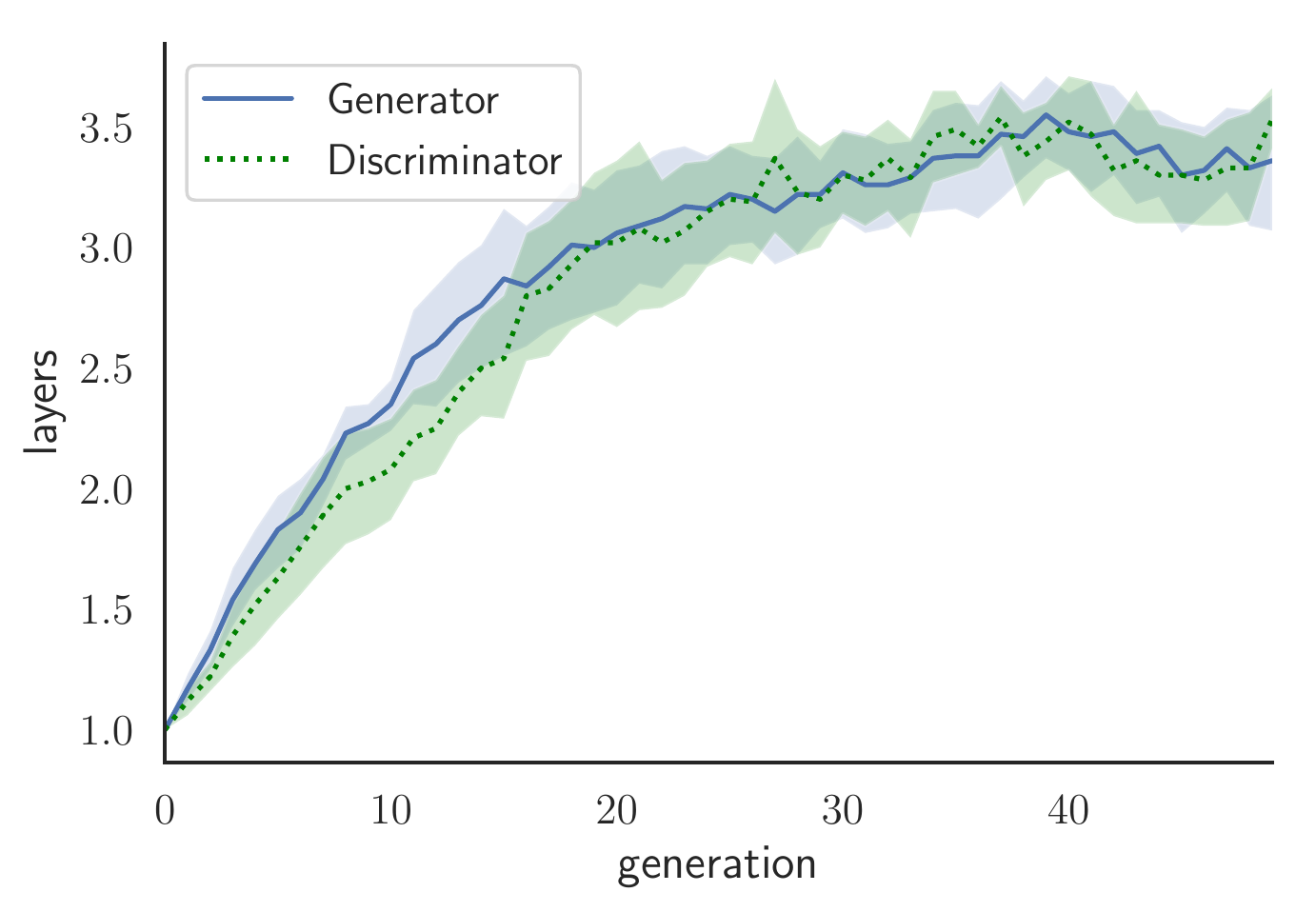}
	\caption{The average number of layers on the Fashion-MNIST dataset.}
	\label{fig:fashionmnist_layers_d}
\end{figure}

\begin{figure}[h]
	\includegraphics[width=0.4\textwidth]{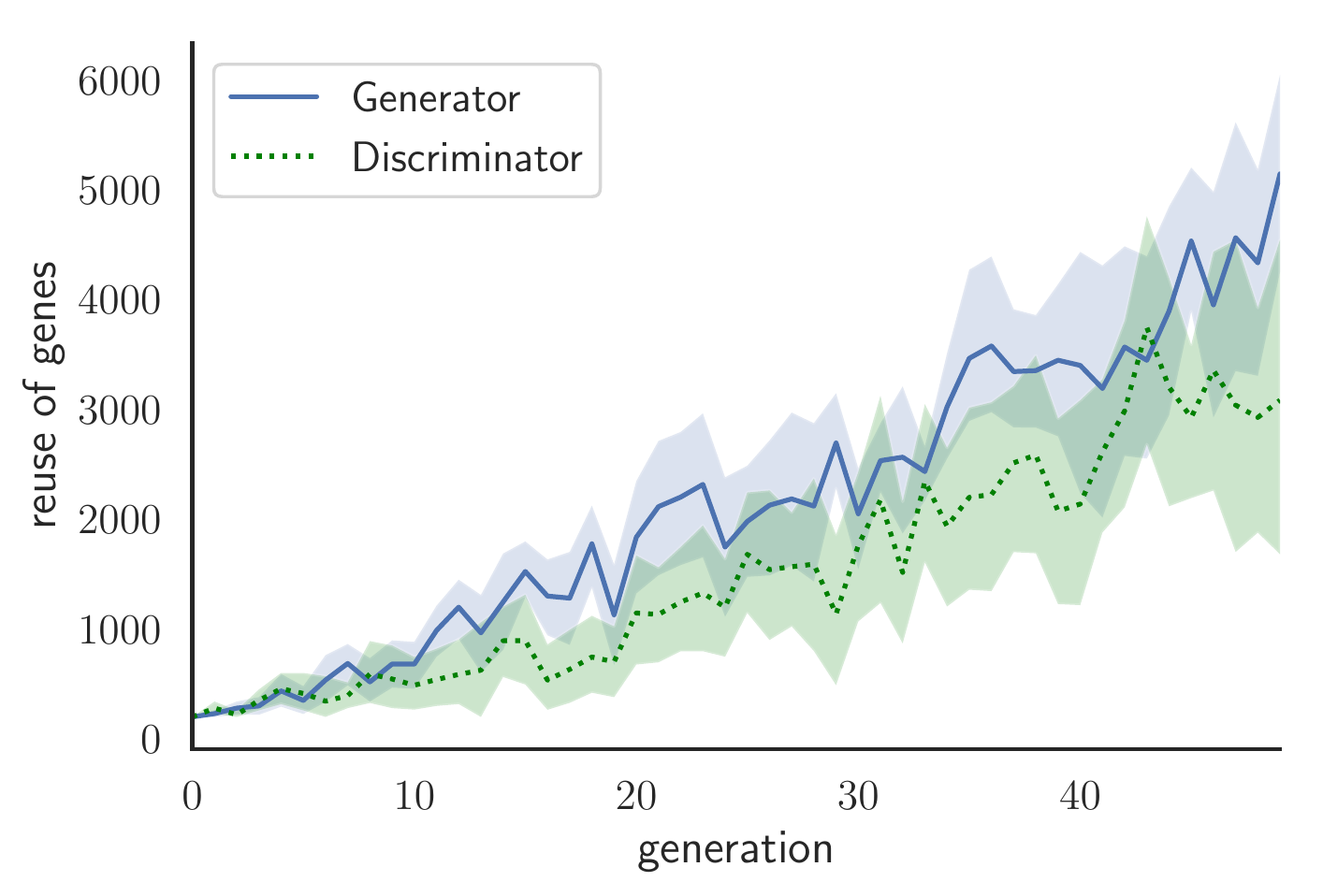}
	\caption{The average number of times a gene was reused when trained on the Fashion-MNIST dataset.}
	\label{fig:fashionmnist_genes_used_d}
\end{figure}

\begin{figure}[h]
	\includegraphics[width=0.4\textwidth]{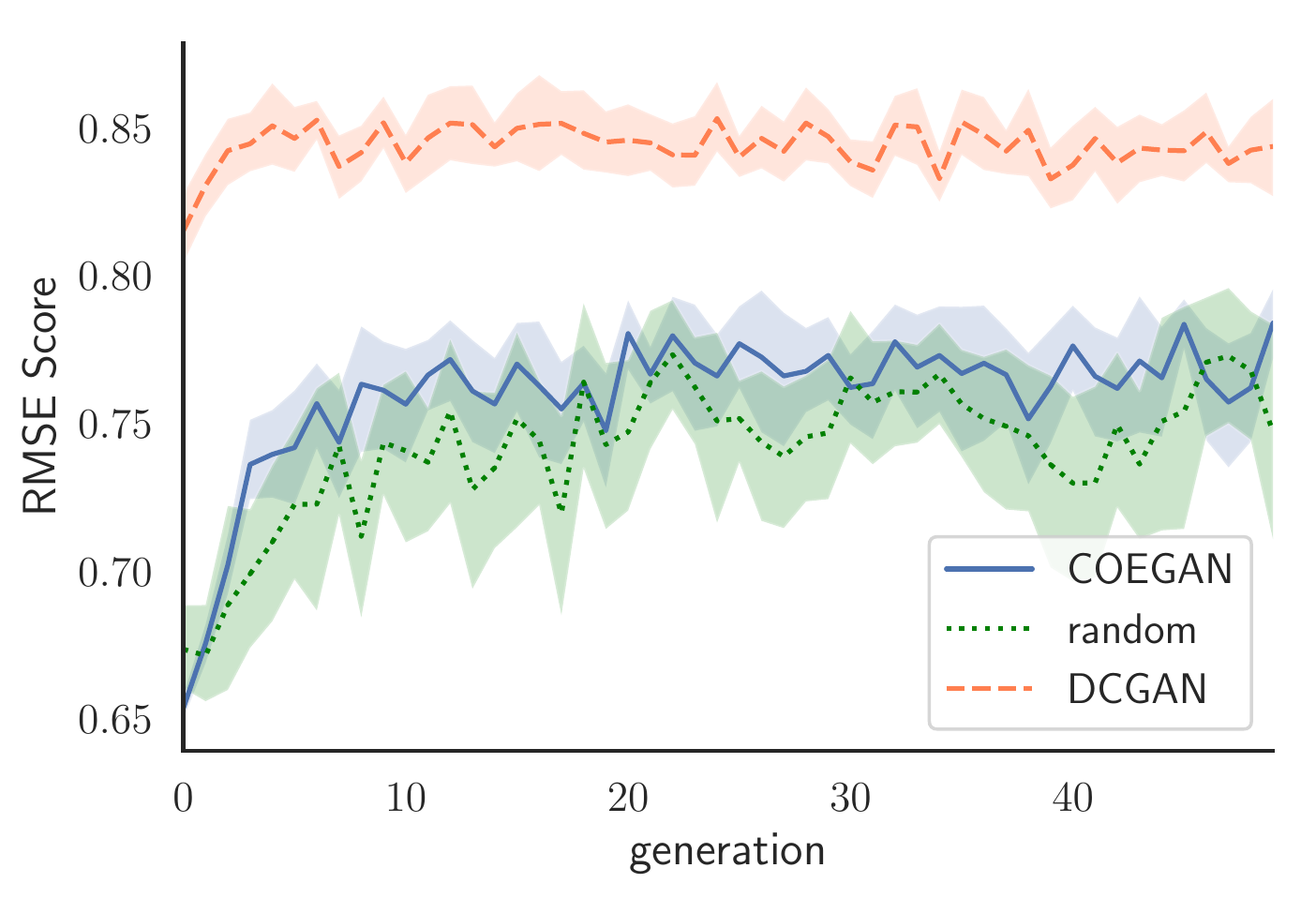}
	\caption{Root mean squared error on the Fashion-MNIST dataset.}
	\label{fig:fashionmnist_rmse_score_g}
\end{figure}

\begin{figure}[h]
	\includegraphics[width=0.4\textwidth]{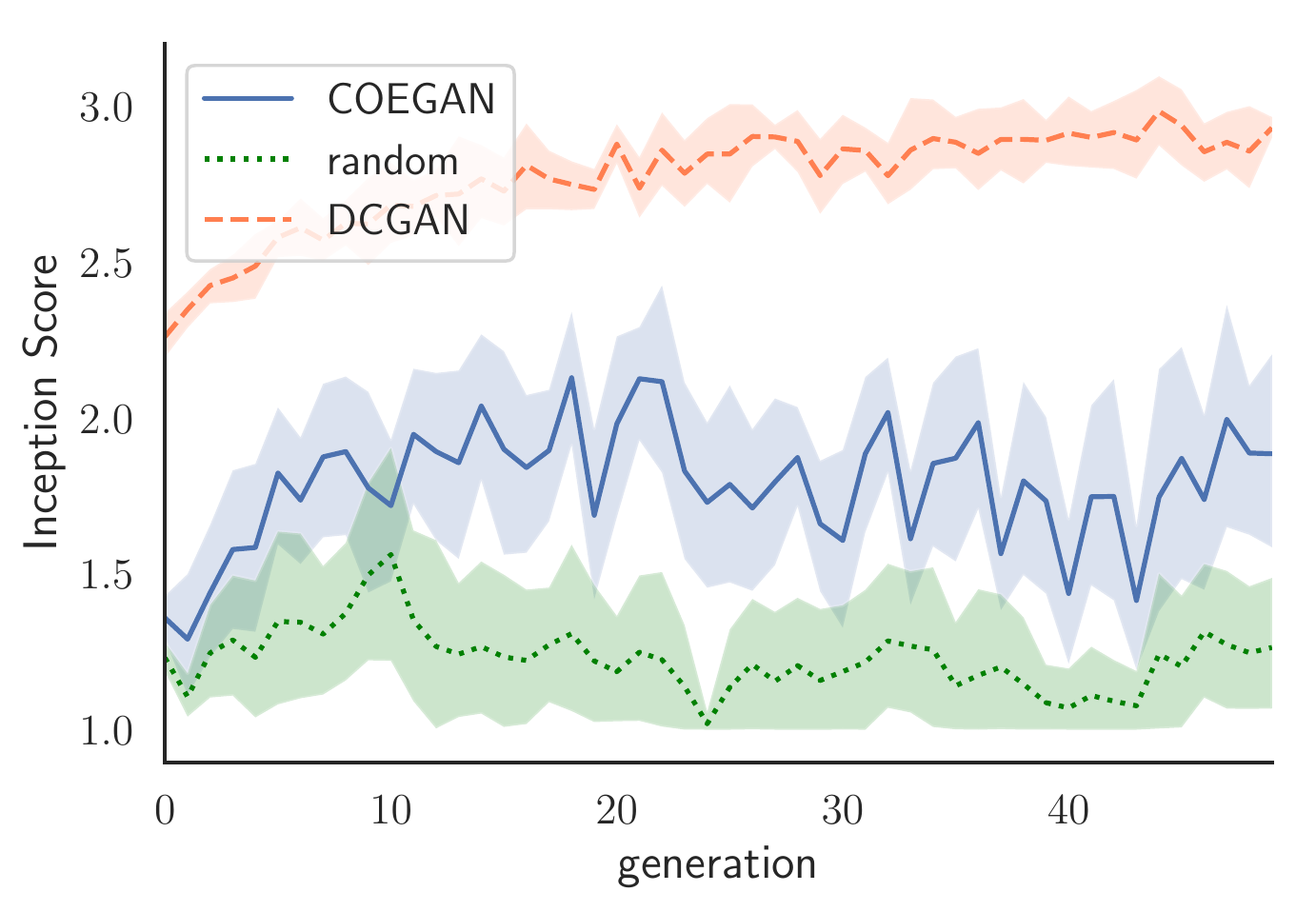}
	\caption{Inception Score on the Fashion-MNIST dataset.}
	\label{fig:fashionmnist_inception_score_g}
\end{figure}

\begin{figure}[h]
	\includegraphics[width=0.4\textwidth]{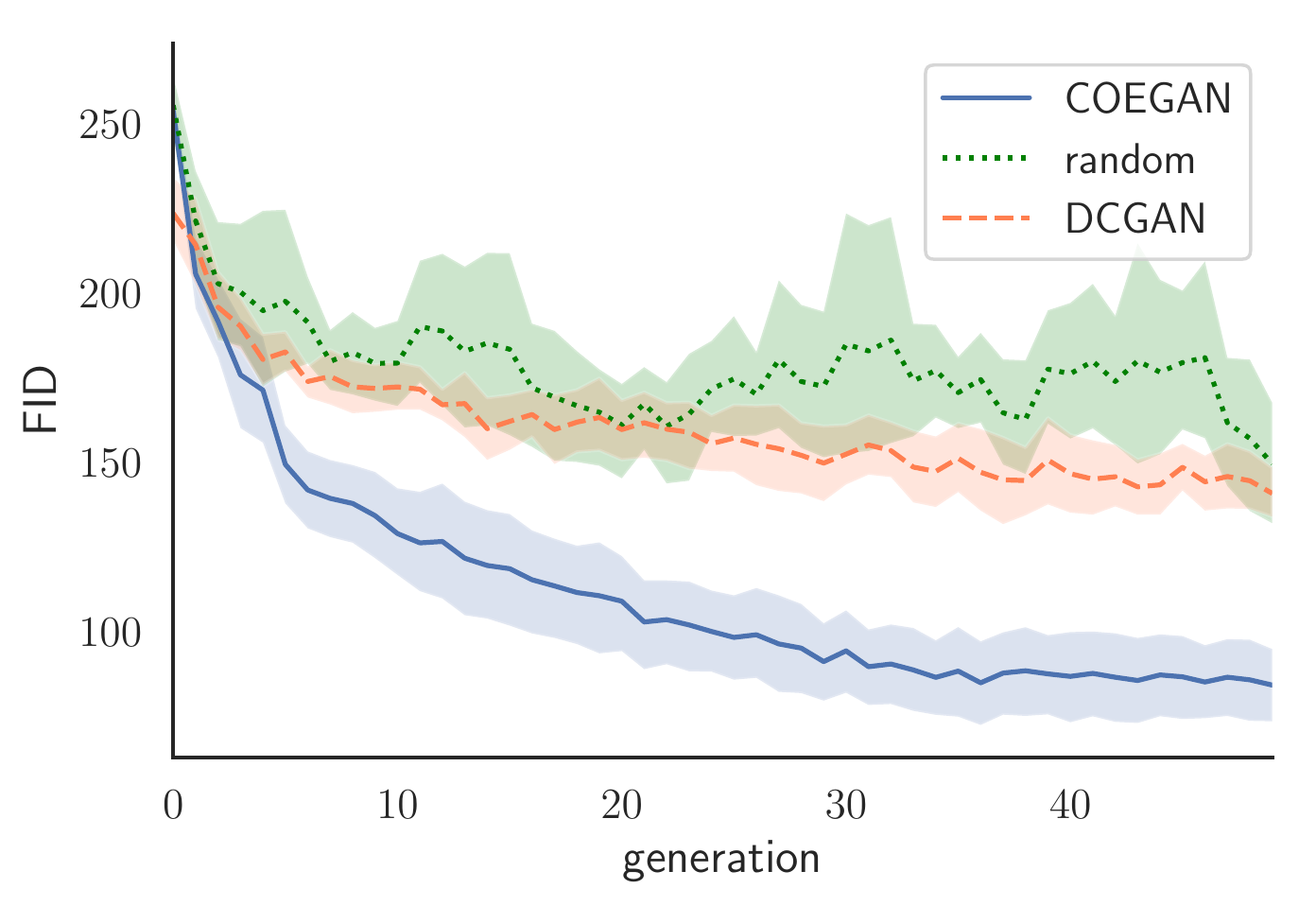}
	\caption{FID Score on the Fashion-MNIST dataset.}
	\label{fig:fashionmnist_fid_score_g}
\end{figure}

The same methodology to assess the performance of COEGAN (used with the MNIST dataset in Section \ref{sec:results_mnist}) was applied with the Fashion-MNIST dataset.
Therefore, Figures \ref{fig:fashionmnist_loss_d}, \ref{fig:fashionmnist_layers_d}, \ref{fig:fashionmnist_genes_used_d}, \ref{fig:fashionmnist_rmse_score_g}, \ref{fig:fashionmnist_inception_score_g} and \ref{fig:fashionmnist_fid_score_g} present results with similar characteristics of the previous results on the MNIST dataset.
As the Fashion-MNIST dataset is slightly more complex than MNIST, we can conclude that our method can be applied in more elaborated datasets.
However, experiments with larger datasets such as CelebA \cite{liu2015deep} and CIFAR-10 \cite{krizhevsky2009learning} should be conducted to support this statement.

As in the MNIST results, we can see in Figure \ref{fig:fashionmnist_fid_score_g} that the FID score for COEGAN outperforms the other methods.
The Inception Score is still better for DCGAN on the Fashion-MNIST dataset.
The progression in the number of layers, presented in Figure \ref{fig:fashionmnist_layers_d} is still similar to the MNIST results.

\begin{figure}[h]
	\includegraphics[width=0.4\textwidth]{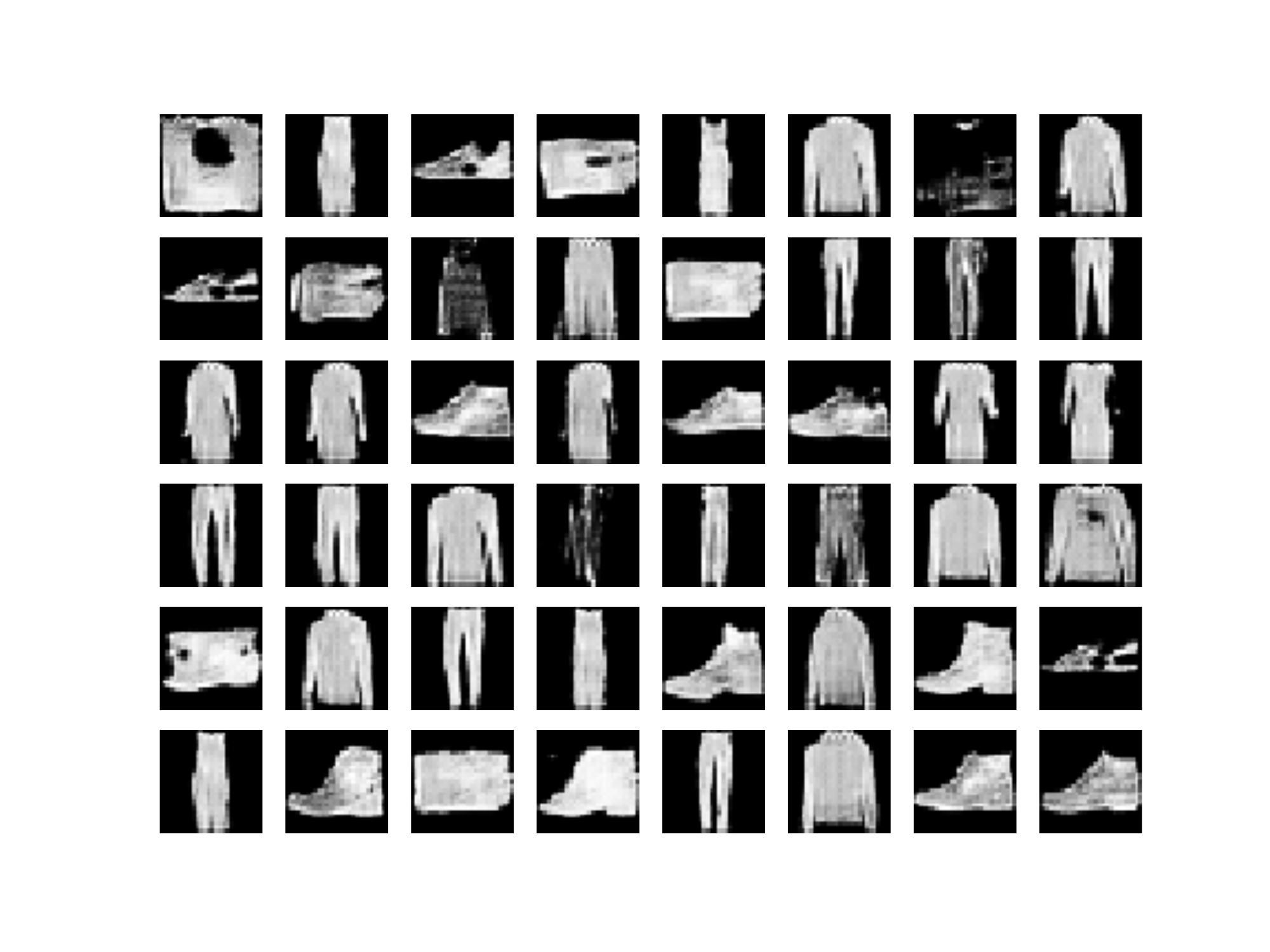}
	\caption{Samples generated by COEGAN when training on the Fashion-MNIST dataset.}
	\label{fig:fashionmnist_samples}
\end{figure}

Figure \ref{fig:fashionmnist_samples} contains samples generated by the best generator found in COEGAN trained with the Fashion-MNIST dataset after $50$ generations.
We can see a variety of images being generated, following the distribution of the input dataset.
As in the results with the MNIST dataset, we also found no evidence of the vanishing gradient and the mode collapse problem in all executions.

\section{Conclusions} \label{sec:conclusions}
Generative adversarial networks (GAN) gained relevance for presenting impressive results in the field of computer vision.
However, stability problems such as the vanishing gradient and the mode collapse problem make the training of a GAN a difficult task.

We present in this paper a model called COEGAN, first proposed in \cite{costa2019coegan}, which uses neuroevolution and coevolution in the coordination of the GAN training process.
COEGAN makes use of the adversarial characteristics of a GAN to apply a coevolution environment.
The model was designed with inspiration on NEAT~\cite{stanley2004competitive} and DeepNeat~\cite{miikkulainen2017evolving}, and also on recent advances in GANs, such as \cite{karras2018progressive}.

In this paper, we presented experiments made with the MNIST and Fashion-MNIST datasets to assess the efficiency of COEGAN.
We found no evidence of the vanishing gradient and the mode collapse problem in all executions of the experiments with COEGAN in both MNIST and Fashion-MNIST datasets.
The selection process and the variation introduced by a diverse population of generators and discriminators contributed to preventing these issues.
Thus, COEGAN presented a more stable training solution than regular GANs.
We compare our results with a random search method and also with a reference architecture based on DCGAN.
The results displayed that COEGAN achieved a FID better than DCGAN and the random method for both datasets.
However, the Inception Score of the DCGAN model was better than COEGAN.
The Inception Score is a metric that has issues to represent the diversity and quality of the samples, being gradually replaced by the FID score in the analysis of the quality of GANs.
We also show that COEGAN is better than a random search model, demonstrating the efficiency of the evolutionary algorithm proposed in this paper.
It is also important to note that COEGAN discovered models for generators and discriminators with less layers than the DCGAN used in our experiments.
Therefore, the evolutionary process that leads to the final models in COEGAN is also relevant to the performance of the models, mainly because of the mechanism of weights transference through generations.

As future works, we will apply the same experiments in larger datasets, such as CelebA \cite{liu2015deep} and CIFAR-10 \cite{krizhevsky2009learning}.
We will also expand the parameters used in the experiments in this paper to enable the generator of larger networks.
Thus, a larger population of generators and discriminators can be used with a larger limit in the number of genes in the genome.

\bibliographystyle{ACM-Reference-Format}
\bibliography{references} 

\end{document}